\documentclass[10pt,journal,compsoc]{IEEEtran}
\usepackage{kotex}
\usepackage{cite}
\usepackage[numbers]{natbib}
\usepackage[pdftex]{graphicx}
\usepackage{xcolor}
\usepackage{tikz}
\usepackage{mathtools}
\usepackage{amsmath}
\usepackage{amsfonts}
\usepackage{bm}

\usepackage{booktabs}
\usepackage{subcaption}

\usepackage{array}
\usepackage{multirow}
\usepackage{url}
\usepackage{hyperref}
\usepackage{algorithmicx}
\usepackage[ruled]{algorithm}
\usepackage{algpseudocode}

\newcommand{\eg}{{\em e.g.}}           
\newcommand{\ie}{{\em i.e.}}           

\newcommand{\rev} {\color{black}}

\begin{document}
%
\title{Medical Transformer: Universal Brain Encoder for 3D MRI Analysis}
%
%
%
%

\author{Eunji Jun,~\IEEEmembership{Student Member,~IEEE,} Seungwoo Jeong, Da-Woon Heo, and~Heung-Il~Suk,~\IEEEmembership{Member,~IEEE}
\IEEEcompsocitemizethanks{\IEEEcompsocthanksitem E. Jun is with the Department of Brain and Cognitive Engineering, Korea University, Seoul 02841, Republic of Korea (e-mail: ejjun92@korea.ac.kr). 
\protect\\
\IEEEcompsocthanksitem S. Jeong and D.-W. Heo are with the Department of Artificial Intelligence, Korea University, Seoul 02841, Republic of Korea (e-mail: sw\_jeong@korea.ac.kr, daheo@korea.ac.kr).
\protect\\
\IEEEcompsocthanksitem H.-I. Suk is with the Department of Artificial Intelligence and the Department of Brain and Cognitive Engineering, Korea University, Seoul 02841, Republic of Korea (e-mail: hisuk@korea.ac.kr).}
}

\IEEEtitleabstractindextext{%
\begin{abstract}
Transfer learning has gained attention in medical image analysis due to limited annotated 3D medical datasets for training data-driven deep learning models in the real world. Existing 3D-based methods have transferred the pre-trained models to downstream tasks, which achieved promising results with only a small number of training samples. However, they demand a massive amount of parameters to train the model for 3D medical imaging. In this work, we propose a novel transfer learning framework, called Medical Transformer, that effectively models 3D volumetric images in the form of a sequence of 2D image slices. To make a high-level representation in 3D-form empowering spatial relations better, we take a multi-view approach that leverages plenty of information from the three planes of 3D volume, while providing parameter-efficient training. For building a source model generally applicable to various tasks, we pre-train the model in a self-supervised learning manner for masked encoding vector prediction as a proxy task, using a large-scale normal, healthy brain magnetic resonance imaging (MRI) dataset. Our pre-trained model is evaluated on three downstream tasks: (i) brain disease diagnosis, (ii) brain age prediction, and (iii) brain tumor segmentation, which are actively studied in brain MRI research. The experimental results show that our Medical Transformer outperforms the state-of-the-art transfer learning methods, efficiently reducing the number of parameters up to about 92\% for classification and regression tasks, and 97\% for segmentation, and still performs well in the scenario where only partial training samples are used. 
\end{abstract}

}

\maketitle

\IEEEdisplaynontitleabstractindextext

%
\IEEEpeerreviewmaketitle

\IEEEraisesectionheading{\section{Introduction}\label{sec:introduction}}
\IEEEPARstart{M}{edical} image analysis has achieved remarkable progresses with the use of deep learning for the past few years \cite{litjens2017survey}. However, building a large 3D dataset for training deep learning models is challenging in the field of medical imaging, due to difficulty of data acquisition and annotation. The lack of annotated data has prompted the development of transfer learning beyond the traditional supervised learning, which allows model training with a small-scale available dataset.

For the transfer learning in medical imaging, the standard approach was to pre-train the existing 2D-based deep neural architectures, \eg, ResNet \cite{he2016deep} and DenseNet \cite{huang2017densely}, using a large-scale natural image datasets such as ImageNet, and then fine-tune the model on small-sized medical imaging data \cite{yu2018recurrent,han2017automatic}. This approach yields better performance than the random initial-based training strategy, especially when a small set of samples are available, but the 3D volume data was split into 2D slices from three planes and pre-trained by 2D models without recovering to 3D-form representations, which inevitably caused the model to lose the 3D spatial information. Accordingly, \cite{ardila2019end} resolved this issue by exploiting the 3D public available models, \eg, recurrent neural networks, trained from the 3D Kinetics dataset \cite{Carreira}.

However, despite the same 3D structure, it is worth emphasizing that the medical dataset of interest is very different to the natural images benchmark datasets in computer vision. In the case of 3D medical data, all scans are registered, resulting into sophisticated yet recurrent patterns. Learning a 3D medical image network transferred from a natural scene video without considering this characteristic leads to a strong bias in medical image analysis. Thus, it motivates the need to learn a source model for transfer learning by directly utilizing the medical imaging datasets capable of learning subtle differences among medical datasets.

To build a source model directly from 3D medical volume data, recent works pre-trained their models for target tasks such as brain parsing and organ segmentation over labeled datasets \cite{Gibson,Chen}. But they were pre-trained for specific applications with a small set of data, thus not well suited for the source model. Meanwhile, \cite{Chen} built a large-scale dataset from a collection of eight annotated medical datasets, and pre-trained the 3D residual network for the segmentation task. Although these fully-supervised approaches may yield more powerful target models, they demanded a volume of annotation efforts to obtain the source model.

Accordingly, many works have adopted a self-supervised pre-training to learn image representations from unlabeled medical imaging data \cite{tajbakhsh2019surrogate,zhuang2019self,zhu2020rubik,Taleb,Zhou}. For example, \cite{zhuang2019self,zhu2020rubik} introduced a proxy task for 3D representation learning by recovering the rearranged and rotated Rubik's cube. In addition, \cite{Taleb} proposed a set of five 3D self-supervised learning (SSL) tasks that extend the existing 2D SSL tasks to 3D form for medical imaging. More recently, \cite{Zhou} devised an image restoration task from a mixture of transformations to pre-train the model via self-supervision without a labeled dataset. However, despite self-supervision, the existing 3D-based transfer learning frameworks require too many learnable parameters to practically apply to high-dimensional medical datasets.

\begin{figure*}[thp!]
\centering
\includegraphics[width=\textwidth]{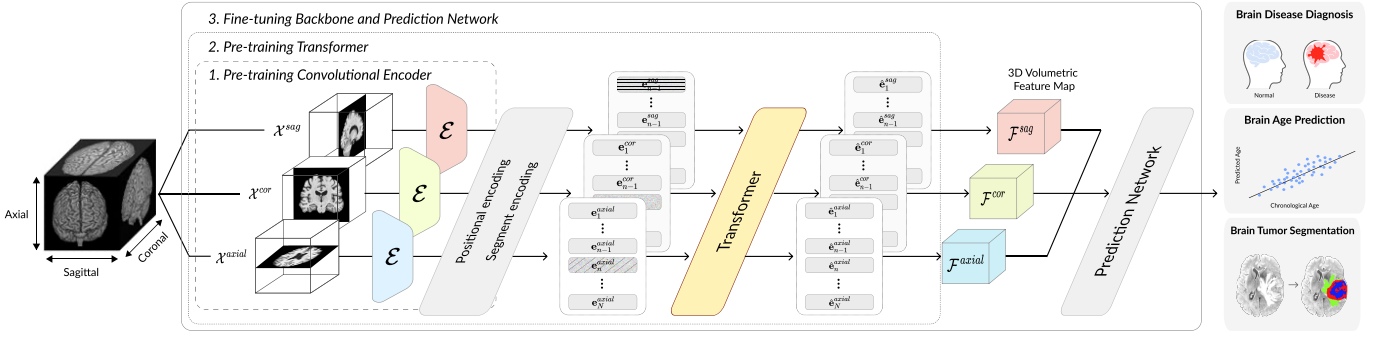}
\caption{Schematic diagram of the proposed Medical Transformer. Based on a multi-view approach, a given 3D volumetric image is split into 2D slices from three planes (sagittal, coronal, axial), and these 2D image slices are fed to the network as inputs. First of all, we pre-train a backbone network that consists of a convolutional encoder and a transformer in a self-supervised learning scheme. Then, after passing through the pre-trained backbone network, the 2D slice features are recovered by their combinations into 3D-form representations, and finally fed into the prediction network for three medical imaging tasks.} 
\label{fig:architecture}
\end{figure*}

In this work, we propose a novel transfer learning framework, called \emph{Medical Transformer}, that effectively models 3D volumetric images in the form of a sequence of 2D image slices. To make a high-level representation in 3D-form empowering spatial relations better, we take a multi-view approach that leverages plenty of information from the three planes of 3D volume, while providing a parameter-efficient training. Specifically, a given 3D volumetric image is split into 2D slices from three planes (sagittal, coronal, axial), and these 2D image slices are fed to the network as inputs. First of all, we pre-train a backbone network consisting of a convolutional encoder and a transformer for masked encoding vector prediction that generates the encoding vectors of the randomly masked images, as our SSL proxy task. Then, after passing through the pre-trained backbone network, the 2D slice features are recovered by their combinations into 3D-form representations, and finally fed into the prediction network for downstream tasks. To validate the effectiveness of our Medical Transformer, we evaluated our pre-trained model on the three target tasks, \ie, brain disease diagnosis, brain age prediction, and brain tumor segmentation which are actively studied in brain MRI research, and achieved remarkable performances for all three tasks in comparison to the respective state-of-the-art (SOTA) models.

\section{Related Work}
\label{sec:relatedwork}

\subsection{Transformers in Vision Models:} 
Recently, many works have adopted transformers in vision models. Among them, Vision Transformer (ViT) \cite{dosovitskiy2020image} divided an image into $16\times 16$ patches, and fed these patches into a transformer \cite{vaswani2017attention} by treating them as tokens. Despite the simple application of a standard transformer, it showed a comparable performance to the convolutional variants by capturing dense and repeatable patterns. However, ViT has a fetal drawback that requires an enormous amount of computation cost and dataset to outperform the competing convolutional methods.

Our approach is similar to ViT in the sense that we use the transformer as a means of dividing image into subsets and modeling their dependencies. However, to take advantage of both convolutional encoder and transformer, the convolutional encoder extracts high-level spatial features, and the transformer extracts their relational features taking into account inter-slice dependencies. In addition, the spatial attention of transformer focuses on important slices, instead of treating each image slice equally, which yields superior performances over 3D-based competing convolutional methods.

\subsection{Fully Supervised Pre-training}
Due to few annotated datasets in the medical imaging domain, modern approaches have used the pre-trained model using ImageNet dataset containing over 14 million annotated images \cite{shin2016deep,tajbakhsh2016convolutional}. Practically, their pre-trained weights from 2D models, such as ResNet \cite{he2016deep} and DenseNet \cite{huang2017densely}, are fine-tuned on medical image analyses, including  skin cancer identification \cite{esteva2017dermatologist}, Alzheimer's disease (AD) diagnosis \cite{ding2019deep}, and pulmonary embolism detection \cite{tajbakhsh2019computer}. But this strategy was limited to apply for 3D medical imaging modalities, \eg, computed tomography (CT) and MRI, inevitably losing 3D anatomical information. Accordingly, in order to capture 3D spatial information, \cite{ardila2019end} utilized the Inflated 3D (I3D) \cite{Carreira} which is trained from the 3D Kinetics dataset, as a feature extractor. However, despite the same 3D structure, there exists a large domain gap between temporal video data of natural scenes and medical data, which leads to improper modeling of medical contexts. In order to alleviate these limitations, recent works have tried to pre-train a source model directly using 3D medical volume data. For example, NiftyNet \cite{Gibson} publicized the pre-trained models, \ie, model zoo, for specific applications such as brain parcellation and organ segmentation. But they were trained with a small-scale dataset, not suitable as a source model for transfer learning. For training the source model from a large-scale dataset, \cite{Chen} aggregated eight annotated medical datasets, and employed them to pre-train a 3D residual network for a segmentation task.

These aforementioned fully supervised pre-training approaches demand massive, high-quality annotated datasets to obtain the source models for transfer learning. On the other hand, our proposed method utilizes a self-supervised learning framework without the need of labeled datasets for 3D medical image analysis.

\subsection{Self-supervised Pre-training:} 
Self-supervised learning has recently gained attention in the field of medical imaging, aiming at learning image representation from unlabeled data \cite{tajbakhsh2019surrogate,zhuang2019self,zhu2020rubik,Taleb,Zhou}. The key challenge for self-supervised learning is to define a suitable proxy task from the unlabeled dataset. For example, the proxy tasks in the computer vision include colorization that recovers gray-scale image to a colored image and {\rev image restoration that restores the shuffled small regions within an image to their original position in the image}. In the meantime, in the medical imaging domain, \cite{zhuang2019self,zhu2020rubik} introduced a proxy task of 3D representation learning by recovering the re-arranged and rotated Rubik's cube. \cite{Taleb} proposed a set of five 3D self-supervised tasks for medical Imaging such as contrastive predictive coding, rotation prediction, jigsaw puzzles, relative patch location, and exemplar networks. \cite{Zhou} has developed a transfer learning framework, called Model Genesis, to learn general image representation by recovering the original sub-volumes of images from their transformed ones derived from non-linear transformation, local shuffling, outer-cutout, and inner cutout.

Our approach differs from the previous self-supervised approaches in obtaining the encoding vector in the representation space, {\rev while many related works utilized an image restoration for a proxy task that recovers the transformed images in an input space}. Our masked encoding vector prediction of the proposed method can effectively learn universal representations. Additionally, compared to the 3D-based competing methods that require a huge amount of learnable parameters, our method provides a parameter-efficient transfer learning framework, still showing superior performances for three medical imaging tasks.


\section{Proposed Method}
\label{sec:proposed_method}
The motivation of our work is to transfer a backbone network pre-trained with a large-scale 3D brain MRI dataset to downstream tasks of 3D MRI analyses. In this section, we illustrate the overall framework of our Medical Transformer as shown in Figure \ref{fig:architecture}. For learning universal representations of 3D brain MRI, we first build a large-scale 3D brain MRI dataset of normal, healthy subjects by collecting three publicly available datasets, \ie, Information eXtraction from Images (IXI)\footnote{\url{https://brain-development.org/ixi-dataset/}}, Cambridge Centre for Ageing and Neuroscience (Cam-CAN)\footnote{\url{https://www.cam-can.org/}}, and Autism Brain Imaging Data Exchange (ABIDE)\footnote{\url{http://fcon_1000.projects.nitrc.org/indi/abide/}}. To handle the data variation problems, \ie., domain shift caused by data acquisition from multi-centers, we conduct {\rev spatial and intensity distribution normalization}.

Based on a \emph{backbone network} including convolutional encoder and transformer and a \emph{prediction network}, our Medical Transformer takes a multi-view approach, in which a given 3D volumetric image is split into 2D slices from three planes (sagittal, coronal, axial), and these 2D image slices are fed to the network as inputs. For transfer learning, it performs the following three steps: (i) pre-training convolutional encoder with an exemplar network, (ii) pre-training transformer for a masked encoding vector prediction as our proxy task of a self-supervised learning (SSL), and (iii) fine-tuning the backbone and prediction network for downstream medical tasks.

\subsection{Notations}
\label{subsec:notations}
In this work, we employ a multi-view approach that regards a 3D volume image as 2D slice images from three planes, \ie, coronal, sagittal, and axial. Given a 3D volume MRI image $\mathcal{X}^{pl}$ (${pl}\in\{\text{cor},\text{sag},\text{axial}\}$), we denote a set of 2D slice images for each plane, $\mathcal{X}^{pl}=\{\mathbf{X}_1^{pl},...,\mathbf{X}_{N_{pl}}^{pl}\}$ ($\mathcal{X}^{pl}\in\mathbb{R}^{w\times d\times h}$), where $N_{pl}$ is the number of slices for the corresponding plane, and $w, d, h$ present width, depth, and height of 3D volume image, respectively. The 2D slice images are fed into the convolutional encoder, leading to a set of embedding vectors $\mathbf{Z}^{pl}=\{\mathbf{z}_1^{pl},...,\mathbf{z}_{N_{pl}}^{pl}\}$  ($\mathbf{z}_{n}^{pl}\in\mathbb{R}^{d_{\text{emb}}}$), where $d_{\text{emb}}$ denotes dimension of an embedding vector. Then, we conduct a positional and segment encoding step before entering the transformer, and obtain the encoding vectors, $\mathbf{E}^{pl}=\{\mathbf{e}_1^{pl},...,\mathbf{e}_{N_{pl}}^{pl}\}$ ($\mathbf{e}_{n}^{pl}\in\mathbb{R}^{d_{\text{emb}}}$).

\subsection{Medical Transformer Network}
\label{subsec:medicaltransformer}
Our Medical Transformer basically consists of a \emph{backbone network} to be pre-trained by a SSL proxy task, and a \emph{prediction network} based on linear fully connected layers for final prediction. For the backbone network, we use a convolutional encoder and a transformer, in which the convolutional encoder extracts the high-level spatial features of the 2D slices for each plane, and then the transformer extracts their relational features by modeling the inter-slice dependencies via attention mechanism, enabling to capture the dependencies of neighboring and distant slices. Here, we employ an independent convolutional encoder for each plane that does not share model parameters as the 3D MRI consists of three planes, \ie, axial, coronal, and sagittal, that have different views for each plane. Then, the resulting spatial features go through positional encoding for ordering slices, and segment encoding for discerning planes before being fed into the transformer. After passing through the transformer, described in detail below, a 3D volume feature map is formed by a combination of slice-wise features for all planes. Our multi-view approach of integrating slice-wise features from all planes efficiently reduces the model parameters, still sufficiently capturing plenty of 3D volumetric representations. Finally, the prediction network outputs the final task-specific prediction. Depending on the downstream medical tasks, we take a multi-scale approach to capture both low- and high-level features for segmentation task, and a single-scale approach to capture high-level features for classification and regression tasks. 

\begin{figure}[tp!]
\centering
\includegraphics[width=0.48\textwidth]{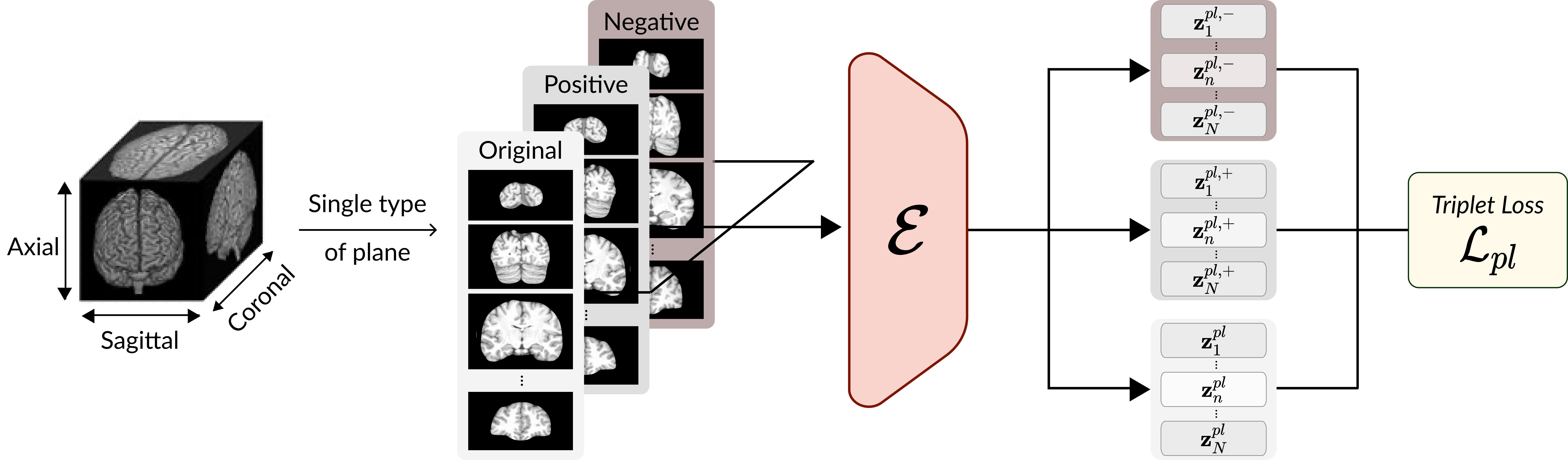}
\caption{\rev Illustration of pre-training a convolutional encoder.} 
\label{fig:pretraining}
\end{figure}

\begin{figure*}[h!]
\centering
\includegraphics[width=\textwidth]{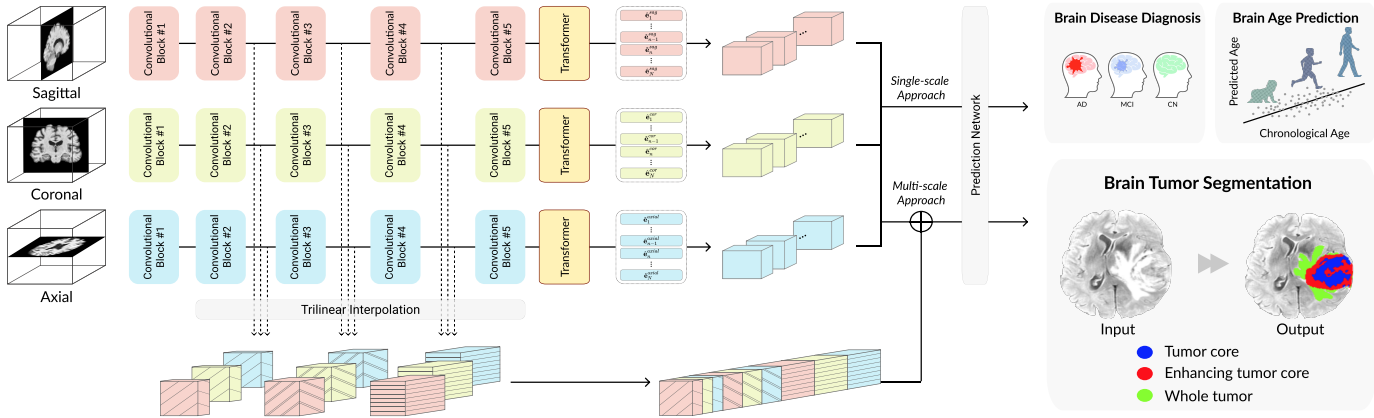}
\caption{Illustration of fine-tuning phase.} 
\label{fig:finetuning}
\end{figure*}

\subsection{Transfer Learning}
\subsubsection{Pre-training Medical Transformer}
\label{subsec:pretraining_resnet}
To make convolutional encoder output spatially meaningful embedding features, we first pre-train the convolutional encoder in a self-supervised learning manner, as shown in Figure \ref{fig:pretraining}. To this end, we use an exemplar network \cite{Dosovitskiy} to derive supervision labels, {\rev for which it} relies on a non-linear transformation for learning appearances as one of the image augmentation techniques. The absolute or relative intensity values of medical images deliver crucial information about the image structure. {\rev To preserve intensity of anatomies during image transformation, we use a Bezier Curve as a smooth and monotonous transformation, which assigns every pixel a unique value, ensuring a one-to-one mapping.}

For training the exemplar network, the triplet loss \cite{wang2015unsupervised} is employed. For each plane, all slice-wise features are average-pooled, and compared with positive pairs (transformed features) and negative pairs (features of samples from other batches) to calculate the triplet loss $\mathcal{L}_{pl}$ as follows:
\begin{equation}
		\mathcal{L}_{pl}=\frac{1}{N_T}\sum_{i=1}^{N_T}\text{max}\{0,D(\tilde{\mathbf{z}}_{(i)}^{pl},\tilde{\mathbf{z}}_{(i)}^{pl,+})-D(\tilde{\mathbf{z}}_{(i)}^{pl},\tilde{\mathbf{z}}_{(i)}^{pl,-})+\alpha\}
\end{equation}
where $\tilde{\mathbf{z}}_{(i)}^{pl}$ is an embedding vector of random training sample that is averaged over slices, $\tilde{\mathbf{z}}_{(i)}^{pl,+}$ is that of transformed version (positive example), $\tilde{\mathbf{z}}_{(i)}^{pl,-}$ is that of different sample from the dataset (negative example), $N_T$ is the number of training samples, $\alpha$ is a margin between positive and negative sample, and $D$ is a function for distance measure, \eg, L2 distance. The final loss is defined by averaging all the losses from the three planes.

Motivated by a successful adoption of the masked language modeling as a proxy task of SSL in BERT \cite{devlin2018bert}, we propose to learn universal 3D MRI representations for a masked encoding vector prediction as our SSL proxy task. Specifically, we regard a slice-wise encoding vector $\mathbf{e}_{n}^{pl}$ as a token, and a set of encoding vectors $\mathbf{E}^{pl}=\{\mathbf{e}_1^{pl},...,\mathbf{e}_{N_{pl}}^{pl}\}$ \ie, an image-level feature, as a sentence. We randomly mask out some input slice images for each plane, and predict the corresponding encoding vectors of the masked images from encoding vectors of the remaining non-masked slices. Here, the convolutional encoder pre-trained in the previous step is frozen, and only the transformer is trained for the masked encoding vector prediction task, so that the transformer can predict the masked encoding vector by figuring out the inter-slice dependencies.  

For modeling inter-slice dependencies, we employ a standard transformer over slice-wise encoding features $\mathbf{E}^{pl}\in\mathbb{R}^{N_{pl}\times d_{\text{emb}}}$. The mathematical formulations are as follows:
\begin{gather}
	\bar{\mathbf{E}}^{pl}=\mathbf{E}^{pl}+\operatorname{Softmax}\left(\frac{\mathbf{E}^{pl}(\mathbf{E}^{pl})^{\top}}{\sqrt{d_{\text{emb}}}}\right)\mathbf{E}^{pl}\\
	\hat{\mathbf{E}}^{pl}=\bar{\mathbf{E}}^{pl}+\sigma(\bar{\mathbf{E}}^{pl}\mathbf{F}_1)\mathbf{F}_2
\end{gather}
where $\sigma(\cdot)$ is a ReLU function for non-linearity, and $\mathbf{F}_1\in\mathbb{R}^{d_{\text{emb}}\times d_{ff}}$ and $\mathbf{F}_2\in\mathbb{R}^{d_{ff}\times d_{\text{emb}}}$ denote two point-wise feed-forward convolutions.

\subsubsection{Fine-tuning Medical Transformer}
The proposed Medical Transformer aims to build the backbone network that works as a universal brain encoder generally transferable for various downstream medical tasks. In order to validate the effectiveness of our pre-trained backbone network, we fine-tune the backbone and prediction network for downstream tasks. To this end, we take two different approaches, \ie, a \emph{single-scale approach} to capture high-level features for classification and regression tasks, and a \emph{multi-scale approach} to capture both low- and high-level features for a segmentation task. Specifically, the multi-scale approach integrates all outputs from the initial convolution layers in the encoder. For instance as shown in Figure \ref{fig:finetuning}, both low- and high-level features from the convolutional blocks, \#2 to \#4, are taken into account, which empowers the representations for the segmentation task.


After passing through the backbone network, the encoding vectors $\hat{\mathbf{E}}^{pl}\in\mathbb{R}^{N_{pl}\times d_{\text{emb}}}$ for each plane construct a 3D volume feature map $\mathcal{F}^{pl}\in\mathbb{R}^{(w\times d\times h) \times d_{\text{emb}}}$. In the single-scale approach, feature maps for all planes are integrated into a final 3D volume feature map $\mathcal{F}\in\mathbb{R}^{(w\times d\times h) \times 3d_{\text{emb}}}$ via concatenation, and in the multi-scale approach, the feature maps from the preceding convolutional blocks are additionally concatenated to the aforementioned final feature map, which is fed into the prediction network. The prediction network consists of a linear fully connected (FC) layer for segmentation task, and a linear FC layer followed average pooling for classification and regression task. During the fine-tuning phase, both the pre-trained backbone and the prediction network are fine-tuned on downstream tasks. 

\begin{table*}[thp!]
\centering
\caption{Performance comparison between competitive methods and the proposed method for three target tasks (brain disease diagnosis (AD/MCI/NC multi-class classification), brain age prediction (regression), and brain tumor segmentation) after cross-validation ($\text{mean}\pm\text{std}$). The entries in bold highlight the best performance among the different approaches for three target tasks.}
\scalebox{0.88}{
\begin{tabular}{l|p{4cm}|l|l|l|l|l}
\toprule 
\multirow{3}{*}{\textbf{Pre-training}} & \multirow{3}{*}{\textbf{Approach}} & \multicolumn{5}{c}{\textbf{Target tasks}} \\ \cline{3-7} 
& & \multicolumn{1}{c|}{Classification} & \multicolumn{1}{c|}{Regression} & \multicolumn{3}{c}{Segmentation} \\ \cline{3-7} 
& & \multicolumn{1}{c|}{mAUC} & \multicolumn{1}{c|}{MAE (years)} & \multicolumn{1}{c|}{Dice WT} & \multicolumn{1}{c|}{Dice TC} & \multicolumn{1}{c}{Dice ET} \\ \toprule
No                                 
& Training from scratch 	            		& $0.7728 \pm 0.0077$	& $4.4357 \pm 0.3260$	    & $0.8649 \pm 0.0094$   & $0.6392 \pm 0.0556$   & $0.4830 \pm 0.0485$ \\ \hline
\multirow{3}{*}{(Fully) supervised} 
& I3D \cite{Carreira}					& $0.7325 \pm 0.0164$   	& $4.6561 \pm 0.3209$		& $0.6607 \pm 0.0542$	& $0.4708 \pm 0.0593$	& $0.0569 \pm 0.0159$	\\ \cline{2-7} 
& NiftyNet \cite{Gibson}				& $0.5031 \pm 0.0165$	& $4.6580 \pm 0.3161$		& $0.8395 \pm 0.0065$   & $0.5295 \pm 0.0148$	& $0.5046 \pm 0.0278$	\\ \cline{2-7} 
& MedicalNet \cite{Chen}				& $0.6910 \pm 0.0063$	& $4.6443 \pm 0.3626$		& $0.7885 \pm 0.0378$	& $0.5681 \pm 0.0572$	& $0.0809 \pm 0.0298$	\\ \hline
\multirow{7}{*}{Self-supervised}

& 3D-RPL \cite{Taleb} 				& $0.4849 \pm 0.0333$   	& $5.1237 \pm 0.7086$		& $0.8555 \pm 0.0462$	& $0.6595 \pm 0.0322$	& $0.3897 \pm 0.0078$	\\ \cline{2-7} 
& 3D-Rotation \cite{Taleb} 			& $0.4965 \pm 0.0077$	& $4.9799 \pm 0.4365$		& $0.8672 \pm 0.0344$	& $0.6756 \pm 0.0204$	& $0.3717 \pm 0.0452$	\\ \cline{2-7} 
& 3D-Jigsaw \cite{Taleb} 				& $0.4950 \pm 0.0202$	& $4.7719 \pm 0.4784$		& $0.8671 \pm 0.0453$	& $0.6739 \pm 0.0218$	& $0.3789 \pm 0.0415$	\\ \cline{2-7} 
& 3D-CPC \cite{Taleb} 				& $0.4943 \pm 0.0109$   	& $5.0091 \pm 0.7856$		& $0.8879 \pm 0.0089$	& $0.6844 \pm 0.0086$	& $0.3760 \pm 0.0159$	\\ \cline{2-7} 
& 3D-Exemplar \cite{Taleb} 			& $0.5085 \pm 0.0163$	& $5.4434 \pm 0.9623$		& $\mathbf{0.8975 \pm 0.0123}$	& $0.6912 \pm 0.0120$	& $0.3819 \pm 0.0134$	\\ \cline{2-7}

& Model Genesis \cite{Zhou} 			& $0.4997 \pm 0.0004$	& $4.6377 \pm 0.3411$		& $0.8505 \pm 0.0203$	& $0.6201 \pm 0.0289$	& $0.0896 \pm 0.0329$	\\ \cline{2-7} 
& Medical Transformer (Ours)			& $\mathbf{0.8347 \pm 0.0072}$	& $\mathbf{3.4924 \pm 0.0863}$		& $0.8733 \pm 0.0086$	& $\mathbf{0.6969 \pm 0.0470}$   & $\mathbf{0.5882 \pm 0.0437}$	\\ \toprule
\end{tabular}
\label{tb:sota_performance}
}
\end{table*}

\section{Experiments}
\label{sec:experiments}
In this section, we evaluated the proposed Medical Transformer for the three medical tasks on publicly available brain MRI datasets. To show the superiority of our proposed method, we compared the results of three tasks with other state-of-the-art methods in the literature, \ie, I3D \cite{Carreira}, NiftyNet \cite{Gibson}, MedicalNet \cite{Chen}, 3D self-supervised methods \cite{Taleb}, and Model Genesis \cite{Zhou}. In addition, we conducted extensive ablation studies for our model to evaluate the effects of different components in the proposed method. In addition to the quantitative validation, we also compared the qualitative results of a segmentation task. For reproducibility of the results, our code is publicly available at ``\url{https://open-after-acceptance}".

\subsection{Experimental Settings}
\subsubsection{Pre-training Medical Transformer}
\label{subsec:experiments_pre-training} 
To build a large-scale brain MRI dataset, we collected T1-weighted structural MRI scans of normal, healthy $1,783$ subjects from IXI (\#=566), Cam-CAN (\#=653) and ABIDE (\#=564) datasets. Only subjects who are 13 years of age or older are considered in ABIDE I and ABIDE II. For pre-processing, we conducted Brain ExTraction (BET), FMRIB Linear Image Registration Toolkit (FLIRT), bias correction, and image intensities min-max normalization. Finally, each 3D MRI volume of  $193\times229\times193$ in size was average-pooled into the size of $96\times114\times96$.

The backbone network comprised the ResNet-18 and the transformer. The network architecture of the transformer was defined with the dimension of an embedding vector ($d_{\text{emb}}$)=16, the dimension of a feed-forward network ($d_{ff}$)=64, the number of heads ($h$)=4, respectively. For the positional encoding, we used a sinusoidal function. We trained our models using the Adam optimizer \cite{DBLP:journals/corr/KingmaB14} with an initial learning rate of $1e-4$ and a multiplicative decay of 0.99 for $150$ epochs (early stopping of 30 patience) using mini-batches of {$10$} samples. We divided samples to five folds, where one fold for the validation set, one fold for the test set, and the remaining folds for the training set, and chose the final optimal model based on the performance over the validation set. For the masked encoding vector prediction task, the masking ratio was set 10\% so that universal representation was effectively learned by reasonable difficulties, \ie, 9, 11, 9 slices are masked for sagittal, coronal, and axial plane, respectively. Pre-training the Medical Transformer was implemented with PyTorch, and trained on GPU NVIDIA GeForce RTX 2080 TI.

\begin{table}[tp!]
\caption{Socio-demographic information summary of ADNI cohort.}
\label{table:demographic}
\scalebox{1}{
\begin{tabular}{l|c|c|c}
\hline
\textbf{Information} & \textbf{CN} & \textbf{MCI} & \textbf{AD} \\ \hline
\#Subjects (Male) & 433 (214) & 748 (447) & 359 (194) \\ \hline
Age (Years) & 74.76 $\pm$ 5.82 & 73.29 $\pm$ 7.57 & 75.29 $\pm$ 7.87 \\ \hline
Education (Years) & 16.29 $\pm$ 2.72 & 15.92 $\pm$ 2.86 & 15.15 $\pm$ 3.05 \\ \hline
MMSE & 29.07 $\pm$ 1.12 & 27.48 $\pm$ 1.82 & 23.21 $\pm$ 2.05 \\ \hline
\end{tabular}
}
\end{table}

\subsubsection{Fine-tuning Medical Transformer}
To validate the effectiveness of our transfer learning framework, we fine-tuned the pre-trained model on several medical tasks, \ie, brain disease diagnosis, brain age prediction, and brain tumor segmentation tasks, which are actively studied in brain MRI researches. The prediction network was built with 1 fully connected hidden layer. With respect to the classification and regression task, the {\rev resulting 3D map after prediction network} was average pooled. We reported the performances of our proposed method and competing methods with the average results from the 5-fold cross validation for all tasks. Fine-tuning our Medical Transformer including the comparative methods was implemented with PyTorch, and trained on GPU NVIDIA GeForce RTX 2080 TI for the classification and regression tasks, and TITAN RTX for the segmentation task.

\textbf{Brain Disease Diagnosis:} Our target task is to differentiate between Alzheimer's disease (AD), mild cognitive impairment (MCI), and cognitively normal (CN), \ie, multi-class classification, which is a crucial, but difficult goal in the study of AD, due to the subtle and diverse morphological changes in the spectrum of AD, MCI, and CN. To this end, we utilized T1-weighted structural MRI scans from the Alzheimer's Disease Neuroimaging Initiative (ADNI) dataset\footnote{\url{http://adni.loni.usc.edu/}}. The demographic information of the dataset is summarized in Table \ref{table:demographic}. The classification performance was evaluated in terms of the multi-class area under the receiver operating curve (mAUC) \cite{hand2001simple}. 

\textbf{Brain Age Prediction:} To verify the versatility of our Medical Transformer on a regression task, we conducted a brain age prediction based on a brain MRI sample, which is regarded as one of the most effective ways for brain aging understanding. T1-weighted structural MRI scans from the ADNI dataset were used same as the classification task, as shown in Table \ref{table:demographic}. The regression results were reported in terms of the mean absolute error (MAE) between the chronological age and the predicted age.

\begin{figure*}[th!]
\centering
\includegraphics[width=1\textwidth]{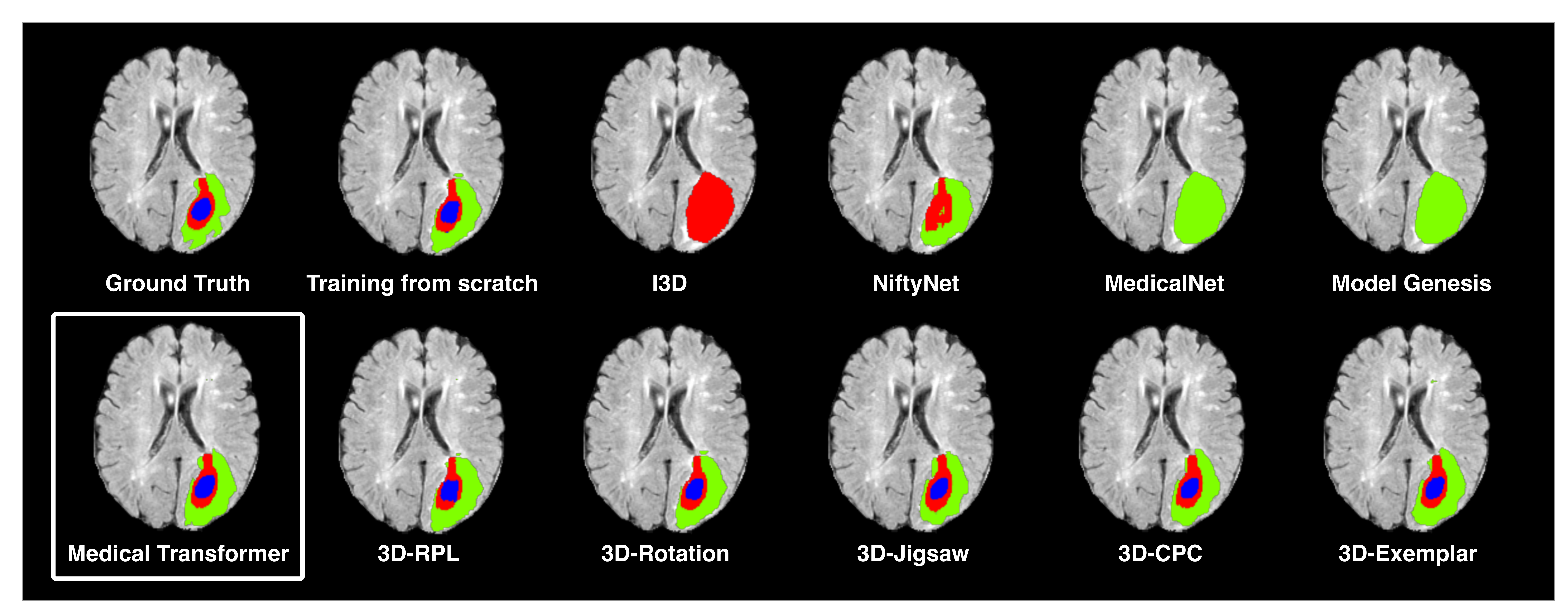}
\caption{Visualization of the segmentation results between our proposed Medical Transformer and the competing methods for comparison of qualitative results. The whole tumor (WT) class includes all visible labels (a union of green, blue and red labels), the tumor core (TC) class if a union of blue and red, and the enhancing tumor core (ET) class is shown in red.} 
\label{fig:qualitative_results}
\end{figure*}

\textbf{Brain Tumor Segmentation} 
To evaluate the proposed method on brain tumor segmentation, we used the Brain Tumor Segmentation (BraTS) 2020 dataset\footnote{\url{https://www.med.upenn.edu/cbica/brats2020/data.html}}. It consists of 369 training samples and 125 validation samples including T1, post-contrast T1-weighted (T1Gd), T2-weighted (T2), and T2-FLAIR. The dataset was pre-processed with skull-striping, interpolation to a uniform isotropic resolution of 1$mm^3$ and registered to the SRI24 space with a dimension of $240\times240\times155$. The original image size is $240\times240\times155$, but the largest crop size of $160\times192\times128$ was used, which ensures that most image content remains within the crop area.

Since the BraTS dataset consists of four different MRI modalities as compared to a single modality in other datasets, there is a difference in the input channel size of the ResNet in the pre-trained backbone network, which causes fine-tuning the backbone network to fail. To handle this issue, {\rev we duplicated the weights of the pre-trained input layer by the number of modalities}. In terms of performance evaluation, we calculated a dice score for each tumor region, \ie, whole tumor (WT), tumor core (TC), and enhancing tumor core (ET).

\begin{table}[thp!]
\centering
\caption{Comparison of the number of model parameters between competitive methods and the proposed method for three target tasks (brain disease diagnosis (AD/MCI/NC multi-class classification), brain age prediction (regression), and brain tumor segmentation). 3D-RPL, 3D-Rotation, 3D-Jigsaw, 3D-CPC and 3D-Exemplar have same architecture, so we refer to it as 3D-SSL. The entries in bold indicate the least number of model parameters among the different approaches.}
\scalebox{0.9}{
\begin{tabular}{p{3.2cm}|c|c|c}
\toprule
\multirow{2}{*}{\textbf{Approach}} & \multicolumn{3}{c}{\textbf{Target tasks}} \\ \cline{2-4} 
 & \multicolumn{1}{c|}{Classification} & \multicolumn{1}{c|}{Regression} & \multicolumn{1}{c}{Segmentation}  \\ \toprule
I3D \cite{Carreira}					& 12.378M & 12.375M  & 17.412M \\ \cline{2-4} 
NiftyNet \cite{Gibson}				& 24.068M   & 24.067M   & 45.614M \\ \cline{2-4} 
MedicalNet \cite{Chen}				& 33.052M   & 33.051M   & 85.798M \\ \hline
3D-SSL \cite{Taleb} 			    & 3.552M	& 3.551M          & 5.649M\\ \cline{2-4}  
Model Genesis \cite{Zhou} 			& 7.093M    & 7.093M    	& 19.076M \\ \cline{2-4}
\textbf{Medical Transformer}		& \textbf{2.402M}    & \textbf{2.401M}    & \textbf{2.410M} 	\\ \toprule
\end{tabular}
\label{tb:parameters}
}
\end{table}

\begin{table*}[th!]
\caption{Performance comparison between supervised learning and transfer learning scheme depending on a ratio of training samples for AD/MCI/NC multi-class classification, brain age prediction, and brain tumor segmentation task. The entries in bold indicate the performance of the transfer learning scheme using a partial dataset that has reached that of the supervised learning scheme using the entire dataset. (mAUC: multi-class area under the receiver operating curve, MAE: mean absolute error, WT: whole tumor, TC: tumor core, ET: enhancing tumor core)}
\centering\scalebox{1.1}{
\begin{tabular}{l|c|c|c|c|c|c}
\toprule
\multirow{2}{*}{\textbf{Scheme}} & \multirow{2}{*}{\textbf{Ratio}}  & \multicolumn{1}{c|}{\textbf{Classification}}  & \multicolumn{1}{c|}{\textbf{Regression}} & \multicolumn{3}{c}{\textbf{Segmentation}} \\ \cline{3-7} 
	&       & \multicolumn{1}{c|}{mAUC} & \multicolumn{1}{c|}{MAE (years)}  &  \multicolumn{1}{c|}{Dice WT} & \multicolumn{1}{c|}{Dice TC}  & \multicolumn{1}{c}{Dice ET} \\ \toprule
\multirow{4}{1.6cm}{Supervised\\Learning}   & 10\%	& $0.5984 \pm 0.0111$   & $4.7720 \pm 0.4769$	& $0.6814 \pm 0.0205$   & $0.1417 \pm 0.1668$   & $0.1710 \pm 0.1537$   \\ \cline{2-7} 
	& 30\%  	& $0.6301 \pm 0.0130$   & $4.5880 \pm 0.3542$   & $0.7921 \pm 0.0125$   & $0.5181 \pm 0.0513$   & $0.5157 \pm 0.0234$   \\ \cline{2-7} 
	& 50\%  	& $0.6749 \pm 0.0166$   & $4.5291 \pm 0.3377$   & $0.8276 \pm 0.0062$   & $0.6509 \pm 0.0415$   & $0.5744 \pm 0.0254$   \\ \cline{2-7} 
	& 70\%  	& $0.6936 \pm 0.0075$   & $4.5055 \pm 0.3515$   & $0.8334 \pm 0.0245$   & $0.6487 \pm 0.0558$   & $0.5641 \pm 0.0355$   \\ \cline{2-7}
	& 100\% 	& $\mathbf{0.7728 \pm 0.0077}$   & $\mathbf{4.4357 \pm 0.3260}$   & $\mathbf{0.8649 \pm 0.0094}$   & $\mathbf{0.6392 \pm 0.0556}$   & $\mathbf{0.4830 \pm 0.0485}$   \\ \toprule
\multirow{4}{1.6cm}{Transfer\\learning} & 10\%	& $0.6112 \pm 0.0288$   & $\mathbf{4.3531 \pm 0.4846}$   & $0.7699 \pm 0.0259$   & $0.4610 \pm 0.1460$   & $\mathbf{0.4221 \pm 0.0906}$\\\cline{2-7} 
	& 30\%  	& $0.7259 \pm 0.0267$   & $4.0142 \pm 0.4403$   & $0.8236 \pm 0.0192$   & $\mathbf{0.6117 \pm 0.0410}$   & $0.5632 \pm 0.0337$   \\ \cline{2-7} 
	& 50\%  	& $\mathbf{0.7729 \pm 0.0132}$   & $3.7819 \pm 0.2780$   & $0.8417 \pm 0.0081$   & $0.6850 \pm 0.0233$   & $0.6216 \pm 0.0110$   \\ \cline{2-7} 
	& 70\%	& $0.7983 \pm 0.0147$   & $3.6834 \pm 0.1955$   & $0.8576 \pm 0.0227$   & $0.6769 \pm 0.0610$   & $0.5389 \pm 0.0488$    \\ \cline{2-7}  
	&100\%  	& $0.8347 \pm 0.0072$   & $3.4924 \pm 0.0863$   & $0.8733 \pm 0.0086$   & $0.6969 \pm 0.0470$   & $0.5882 \pm 0.0437$   \\ \toprule    
\end{tabular}
\label{tb:training_sample_ratio}
}
\end{table*}

\subsection{Results}

\subsubsection{Brain Disease Diagnosis}
Table \ref{tb:sota_performance} compares the results of our proposed method with those of the comparative models for AD, MCI, and NC classification in terms of mAUC. For the competing methods, we took an encoder from the pre-trained model, and appended a fully-connected layer similar to our prediction network. Our Medical Transformer achieved the best classification performance when comparing to 3D-based SOTA methods. Compared to training from scratch that randomly initialized the model weights, adopting the transfer learning scheme showed a clear superiority in performance by taking advantage of the universal brain representations and accordingly better initialization weights in fine-tuning the target task model. Among the comparative methods, I3D demonstrated a competitive performance by utilizing the recurrent layers as a feature extractor, as the classification task benefits from a high level of embedding features. In addition, MedicalNet presents a comparable performance by pre-training the backbone network using a large-scale medical dataset.

These experimental results validated the efficacy of our multi-view approach to extract 3D volumetric characteristics as well as our SSL proxy task to effectively learn universal representations. In addition, pre-training the backbone network using a large corpus MRI dataset attributed to the superior performance of our method.

\subsubsection{Brain Age Prediction}
We evaluated our Medical Transformer with the competing baseline methods on brain age prediction task in Table \ref{tb:sota_performance}. It is noteworthy that our proposed method outperformed all counterparts with a large margin. In particular, while all competing methods showed inferior performance for this regression task, training from the scratch and Medical Transformer achieved remarkable performance with least MAE, supporting the effectiveness of our multi-view approach to make 3D high-level representations from the three planes of a volume.

\subsubsection{Brain Tumor Segmentation}
As shown in Table \ref{tb:sota_performance}, it can be seen that our Medical Transformer achieved the best or comparable performance on the brain tumor segmentation task, specifically comparable performance in the evaluation of Dice WT and TC, and the best in the evaluation of Dice ET, which is the most difficult among tumor types.

The 3D-SSL methods showed consistently better performances in terms of Dice WT and Dice TC, but presented a relatively poor result in Dice ET. This implies that they effectively learn 3D structural representations through 3D SSL proxy tasks, but are inferior to capturing the sophisticated tumor regions. On the contrary, I3D performed the poorest among the comparison methods, due to a large domain gap of transfer learning between the 3D Kinetics dataset and the medical dataset. In addition, MedicalNet showed relatively low performances in all metrics, despite the largest number of parameters among the competing methods.

Our result of obtaining the best performance in the Dice ET evaluation showed the effectiveness of our method for capturing more sophisticated tumor regions by generating 3D volumetric features from a combination of three plane encoding features via our multi-view approach. Furthermore, the multi-scale approach during fine-tuning attributed the performance improvement by integrating low- and high-level features.

Figure \ref{fig:qualitative_results} compares the qualitative results of segmentation between the competing methods and ours. We observed that the segmentation result of the proposed method was similar to that of 3D-SSL, and although it is not possible to accurately capture the border region of the WT, but the three tumor types were better distinguished among the comparison methods. {\rev This result is reasonable as in this work, we focus on proposing a transfer learning framework that is generally applicable to a variety of downstream tasks in a parameter-efficient manner.}

\begin{table}[tp!]
\caption{Performance comparison between w/ and w/o transformer. The entries in bold present better performance between w/o and w/ transformer.}
\label{tb:ablation_transformer}
\scalebox{0.95}{
\begin{tabular}{l|l|c|c}
\toprule
\textbf{Task}                       		& \textbf{Measure}    & \textbf{w/o Transformer}		& \textbf{w/ Transformer} \\ \toprule
Classification                      		& mAUC                	& $0.8204 \pm 0.0200$         	& $\mathbf{0.8347 \pm 0.0072}$ \\ \hline
Regression                          		& MAE (years)         	& $3.6373 \pm 0.2432$         	& $\mathbf{3.4924 \pm 0.0863}$\\ \hline
\multirow{3}{*}{Segmentation}       	& Dice WT             	& $0.8695 \pm 0.0070$         	& $\mathbf{0.8733 \pm 0.0086}$ \\ \cline{2-4}
                                    			& Dice TC             	& $0.6363 \pm 0.0576$         	& $\mathbf{0.6969 \pm 0.0470}$ \\ \cline{2-4}
                                    			& Dice ET             	& $0.5063 \pm 0.0511$         	& $\mathbf{0.5882 \pm 0.0437}$ \\ \toprule
\end{tabular}
}
\end{table}

\subsubsection{Number of Parameters} 
To investigate the power of our parameter-efficient Medical Transformer, we compared the number of model parameters between ours and the comparative methods for three downstream tasks. As shown in Table \ref{tb:parameters}, the proposed method has fewer parameters as much as 92.74 \% for classification, 92.73 \% for regression, and 97.19 \% for segmentation compared to the {\rev comparison method}, but still showing superior performances.

\subsubsection{Ablation Studies} 
First of all, to investigate the effect of transfer learning that performs well with a handful of training samples, we simulated experiments with a fewer labeled data. Specifically, we fine-tuned the model with a partial dataset, \ie, only 10\%, 30\%, 50\%, and 70\% samples of training dataset for three target tasks. Table \ref{tb:training_sample_ratio} compares the results of supervised learning (training from scratch) and transfer learning for each task. The performance of transfer learning consistently showed a significant performance improvement over the supervised learning for the three tasks in all ratio scenarios. Specifically, for brain disease diagnosis, the transfer learning scheme using about 50\% training data already achieved the similar performance to the supervised learning scheme using the entire data, thus reducing nearly 50\% annotation efforts. In addition, using only 30\% of training samples, the performance for brain age prediction was better than that of training from scratch using full samples, which suggests that about 70\% of the annotation cost from supervised learning scheme could be saved by initializing with our Medical Transformer. In terms of brain tumor segmentation, Medical Transformer achieved similar performances to training from scratch by using nearly 70\% on WT, 40\% on TC, and 20\% on ET, where in particular, the effect of reducing annotation efforts was remarkable in the evaluation of ET. Therefore, these experimental results suggest that our Medical Transformer can alleviate the lack of annotated 3D MRI samples through an annotation-efficient transfer learning scheme for a variety of downstream tasks.
	
Furthermore, to examine the effectiveness of using the transformer in our framework, we compared the performance between without and with transformer as shown in Table \ref{tb:ablation_transformer}. In fact, w/ transformer achieved relatively higher performances, compared to w/o transformer for three tasks. Thus, this experimental result suggests that our approach of adopting the transformer helps to capture the volumetric features of 3D MRI by allowing the model to take into account the relations over the neighboring and distant slices.

\section{Conclusion}
\label{sec:conclusion}
In this work, we proposed a novel transfer learning framework, called Medical Transformer, that effectively models 3D volumetric images in the form of a sequence of 2D image slices. To learn high-level 3D volumetric representations, we took a multi-view approach that leveraged plenty of information from the three planes of a volume image, while providing parameter-efficient training. As a result of evaluating our pre-trained model on brain disease diagnosis, brain age prediction, and brain tumor segmentation tasks, our Medical Transformer outperformed the SOTA transfer learning methods, efficiently reducing the number of parameters up to about 92\% for classification and regression tasks, and 97\% for segmentation.

\ifCLASSOPTIONcaptionsoff
  \newpage
\fi



%
%
%

\bibliographystyle{IEEEtran}
\bibliography{egbib}

%


%
%




\end{document}